\documentclass[lettersize,journal]{IEEEtran}
\usepackage{amsmath,amsfonts}
\usepackage{algorithmic}
\usepackage{algorithm}
\usepackage{array}
\usepackage[caption=false,font=normalsize,labelfont=sf,textfont=sf]{subfig}
\usepackage{textcomp}
\usepackage{stfloats}
\usepackage{url}
\usepackage{verbatim}
\usepackage{graphicx}
\usepackage{cite}
\usepackage{amsmath}    
\usepackage{amssymb}    
\usepackage{pifont} 
\usepackage{booktabs} 
\usepackage{tikz,xcolor}
\definecolor{lime}{HTML}{A6CE39}
\DeclareRobustCommand{\orcidicon}{%
    \begin{tikzpicture}
    \draw[lime, fill=lime] (0,0) 
    circle [radius=0.16] 
    node[white] {{\fontfamily{qag}\selectfont \tiny ID}};    \draw[white, fill=white] (-0.0625,0.095) 
    circle [radius=0.007];    \end{tikzpicture}
    \hspace{-2mm}}
\foreach \x in {A, ..., Z}{%
    \expandafter\xdef\csname orcid\x\endcsname{\noexpand\href{https://orcid.org/\csname orcidauthor\x\endcsname}{\noexpand\orcidicon}}
    }

\usepackage{hyperref}
\hypersetup{
    colorlinks = true, 
    linkcolor  = blue,  
    citecolor  = blue,  
    urlcolor   = black   
}

\usepackage{etoolbox}
\makeatletter
\patchcmd{\IEEEbiography}{\vskip 3.5\p@ \@plus 1\p@ \@minus 1\p@}{}{}{}
\newlength{\biographydescwidth}
\makeatother
\begin{document}

\title{UAV-VL-R1: Generalizing Vision-Language Models via Supervised Fine-Tuning and Multi-Stage GRPO for UAV Visual Reasoning}

\author{%
Jiajin Guan\orcidA{},
Haibo Mei\orcidB{},
Bonan Zhang\orcidC{},
Dan Liu\orcidD{},
Yuanshuang Fu\orcidE{},
Yue Zhang\orcidF{}

\thanks{Received 12 June 2025; revised 30 July 2025; accepted 5 August 2025. Date of publication 12 August 2025; date of current version 7 October 2025. This work was supported in part by the National Natural Science Foundation of China under Grant 62206188, and in part by the Fundamental Research Funds for the Central Universities under Grant ZYGX2022YGRH008.(Corresponding author: Haibo Mei.)}%
\thanks{Jiajin Guan, Bonan Zhang, Dan Liu, and Yuanshuang Fu are with the Research Institute of Electronic Science and Technology, University of Electronic Science and Technology of China, Chengdu 611731, China (e-mail: jiajin.guan03@gmail.com; zhangbonan37@gmail.com; liudan@uestc.edu.cn; fyuanshuang01@gmail.com).}%
\thanks{Haibo Mei and Yue Zhang are with the School of Aeronautics and Astronautics, University of Electronic Science and Technology of China, Chengdu 611731, China (e-mail: haibo.mei@uestc.edu.cn; logas2922@outlook.com).}%
\IEEEpubidadjcol
\thanks{This article has open-source code and datasets available at \\
\url{https://github.com/Leke-G/UAV-VL-R1}. \\}%
}




\maketitle

\begin{abstract}
Recent advances in vision-language models (VLMs) have demonstrated strong generalization in natural image tasks. However, their performance often degrades on unmanned aerial vehicle (UAV)-based aerial imagery, which features high resolution, complex spatial semantics, and strict real-time constraints. These challenges limit the applicability of general-purpose VLMs to structured aerial reasoning tasks.
To address these challenges, we propose UAV-VL-R1, a lightweight VLM explicitly designed for aerial visual reasoning. It is trained using a hybrid method that combines supervised fine-tuning (SFT) and multi-stage reinforcement learning (RL). We leverage the group relative policy optimization (GRPO) algorithm to promote structured and interpretable reasoning through rule-guided rewards and intra-group policy alignment.
To support model training and evaluation, we introduce a high-resolution visual question answering dataset named HRVQA-VL, which consists of 50{,}019 annotated samples covering eight UAV-relevant reasoning tasks, including object counting, transportation recognition, and spatial scene inference.
Experimental results show that UAV-VL-R1 achieves a 48.17\% higher zero-shot accuracy than the \texttt{Qwen2-VL-2B-Instruct} baseline and even outperforms its 72B-scale variant, which is 36$\times$ larger, on multiple tasks. Ablation studies reveal that while SFT improves semantic alignment, it may reduce reasoning diversity in mathematical tasks. GRPO-based RL compensates for this limitation by enhancing logical flexibility and the robustness of inference.
Additionally, UAV-VL-R1 requires only 3.9\,GB of memory under FP16 inference and can be quantized to 2.5\,GB with INT8, supporting real-time deployment on resource-constrained UAV platforms.
\end{abstract}


\section{Introduction}
\IEEEPARstart{V}{ision} language models (VLMs) have achieved remarkable progress in reasoning over natural images (e.g., GPT-4o \cite{Gpt-4o}, CogVLM \cite{CogVLM}, and Qwen2-VL \cite{Qwen2-VL}), demonstrating strong performance in tasks such as visual question answering and scene description. These results were achieved through the combined use of Supervised Fine-Tuning (SFT) and Reinforcement Learning (RL)-driven post-training mechanisms, particularly RL, which has demonstrated outstanding performance in enhancing model generalization capabilities. RL has been widely applied in training both LLMs and VLMs \cite{DeepSeek-V2, RL-VLM-F, Vision-R1, Med-R1}, achieving better performance than traditional SFT approaches \cite{OpenAI-o1, Kimi-1.5}.
Despite their success on natural image tasks, mainstream VLMs are primarily trained on generic datasets and struggle to adapt to UAV-captured aerial imagery, which features unique perspectives, high resolutions, and dynamic spatial structures. These models often exhibit poor perception accuracy, weak spatial reasoning, and limited structured output in aerial scenarios. Moreover, UAV applications such as disaster monitoring and visual navigation demand high robustness and real-time performance, where general-purpose VLMs fall short in generalization and deployment efficiency. As a result, developing lightweight VLMs with structured reasoning, strong interpretability, and adaptability to aerial task diversity has become a pressing challenge in low-altitude intelligent perception.

Current VLMs typically rely on SFT as the dominant training paradigm \cite{GPT-4, Reasoning-Tasks}. While effective in closed-domain settings, SFT encourages pattern memorization over reasoning, limiting generalization in structured inference tasks \cite{SFT-RL}. Additionally, most pretraining corpora are based on natural images, which differ significantly from UAV imagery in domain characteristics \cite{remote-sensing}, making it difficult for VLMs to extract meaningful features and spatial relations from a bird’s-eye view. Furthermore, existing models often lack explicit Chain-of-Thought supervision or structured guidance, resulting in uncontrollable and unexplainable black-box outputs. These factors collectively make traditional SFT-based VLMs poorly suited for real-world multimodal transfer and high-complexity reasoning.

\IEEEpubidadjcol
To address these challenges, we present UAV-VL-R1, an SFT and multi-stage RL-enhanced VLM tailored for aerial visual question answering, with a focus on improving interpretability and cross-task generalization. Unlike conventional SFT, which aligns static output patterns, RL actively uses reward-based feedback to explore various reasoning paths, ensuring stable and traceable inference. Recent studies have shown that RL-optimized VLMs outperform SFT-based counterparts under open-domain and weakly defined task boundaries \cite{Fine-Tuning, chen2025r1v}.

Considering the structured nature of aerial imagery, we adopt Group Relative Policy Optimization (GRPO) \cite{grpo} as the core policy update mechanism. Compared to Proximal Policy Optimization (PPO) \cite{ppo} and Direct Preference Optimization (DPO) \cite{dpo}, GRPO estimates intra-group relative advantages, reducing training variance and enhancing policy stability, especially effective for structured reasoning tasks. We further design a dual-objective reward function combining format compliance and answer correctness, encouraging semantically accurate and structurally interpretable outputs. Through this integration of structural guidance and stable policy learning, UAV-VL-R1 achieves robust rule alignment, enhanced reasoning consistency, and low resource overhead, making it well-suited for UAV-based visual reasoning in real-world deployment scenarios. For UAV-specific scenarios with strong spatial regularity and structured semantics, this approach proves particularly beneficial in achieving interpretable and generalizable reasoning.

\begin{figure}[!h]  
\centering
\includegraphics[width=\linewidth]{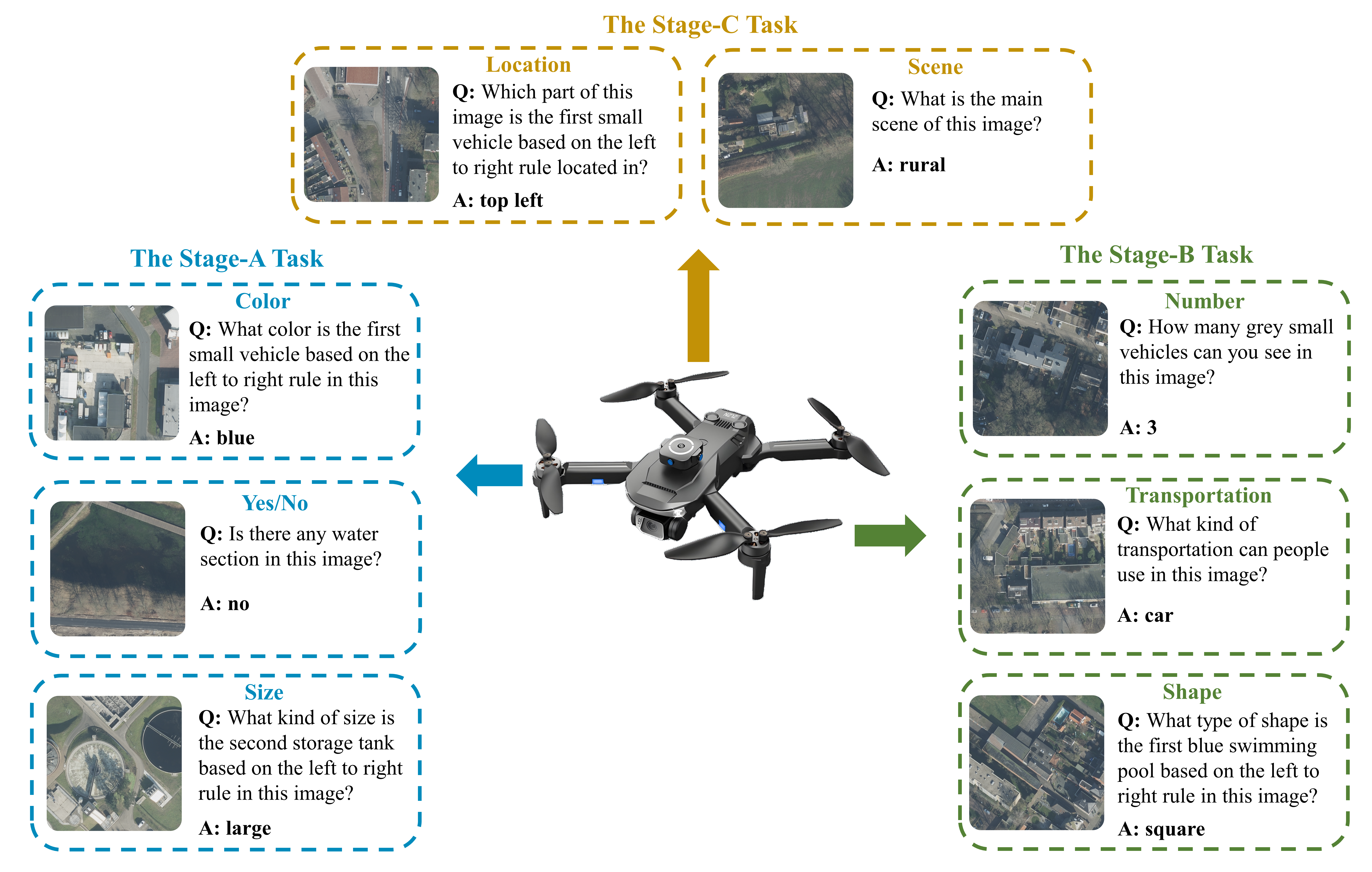}  
\caption{Overview of the HRVQA-VL dataset task structure. We organize tasks into three stages: basic attribute recognition (Stage-A Task), object reasoning and counting (Stage-B Task), and spatial and semantic understanding (Stage-C Task). Each task follows a standardized \textit{image-question-answer} format, enabling structured reasoning and cross-task evaluation under a staged training regime.}
\label{fig:Data_inform}
\end{figure}
To support the training and evaluation of UAV-VL-R1, we construct a high-resolution aerial dataset named HRVQA-VL, built upon real-world UAV imagery and structured into a three-phase training process(see Fig. \ref{fig:Data_inform}).

\begin{itemize}
    \item The Stage-A task covers low-level perceptual tasks such as color, size, and yes/no questions, testing the model’s visual awareness and object-level recognition.
    \item The Stage-B task includes moderately complex tasks like number counting, transportation type, and shape recognition, which require object understanding and relational reasoning.
    \item The Stage-C task addresses high-level semantic reasoning through location inference and scene classification, assessing the model’s capacity to comprehend spatial structure and global image semantics.
\end{itemize}
All samples follow a standardized image–question–answer triplet format. The stage-wise design facilitates multi-stage RL optimization, supports structured reasoning under increasing task complexity, and provides a stable foundation for evaluating the model’s transfer and generalization capabilities in subsequent sections.

In summary, the main contributions of this article are summarized as follows.
\begin{enumerate}
    \item \textbf{Hybrid SFT + Multi-Stage GRPO with Joint Reward Design for UAV Reasoning}: We propose UAV-VL-R1, the first lightweight VLM tailored for eight diverse UAV aerial reasoning tasks. The training framework jointly leverages SFT for semantic alignment and multi-stage GRPO for progressive reasoning enhancement. This process is coupled with a dual-objective rule-based reward function that combines format compliance and answer correctness, forming an integrated optimization strategy that drives the model to produce semantically accurate, structurally consistent, and interpretable reasoning paths.
    
    \item \textbf{Insights into the role of SFT in reinforcement learning}: Through comprehensive ablation studies, we analyze the impact of SFT when applied prior to multi-stage GRPO optimization. Results show that SFT significantly improves early-stage stability and boosts performance on spatial understanding and semantic classification tasks. However, it may hinder numerical reasoning (e.g., counting), suggesting a trade-off between linguistic alignment and reasoning diversity across training paradigms.
    
    \item \textbf{Strong Generalization and Edge Deployment}: UAV-VL-R1 achieves 48.17\% higher multitask accuracy than the base Qwen2-VL-2B-Instruct model and outperforms its SFT-enhanced version by 45.20\%. It even surpasses the 72B Qwen2-VL-Instruct model (72.13\% vs. 46.67\%) while requiring only 3.9 GB (FP16) or 2.5 GB (GPTQ INT8) inference memory \cite{GPTQ}, making it well-suited for on-board UAV deployment.
\end{enumerate}

\section{Related Works}
\subsection{General VLMs and Aerial Image Reasoning VLMs}
In recent years, significant advancements have been achieved in Vision-Language Models (VLMs) for natural image tasks. Models such as Kimi-VL \cite{Kimi-VL} and MiniCPM-V \cite{MiniCPM-V} have demonstrated strong capabilities in visual question answering (VQA) and multimodal reasoning. However, these general-purpose models are predominantly trained on natural image distributions and often exhibit significant performance degradation when applied to aerial imagery, which differs substantially in structural characteristics.Aerial images exhibit high resolution, strong perspective distortion, and complex spatial layouts, resulting in challenges such as extreme scale variation, heavy occlusion, and sparse semantic cues \cite{hrvqa, HazyDet}. These characteristics make tasks like multi-object recognition and spatial reasoning particularly difficult.

Meanwhile, most existing approaches rely heavily on supervised fine-tuning (SFT) or contrastive learning paradigms \cite{Remoteclip, RSVLM}, which, while effective on closed-set benchmarks, often fall short in terms of task generalization and explicit reasoning capabilities \cite{SFT-or-RL}.
To tackle these challenges, we propose UAV-VL-R1, a lightweight VLM built on Qwen2-VL-2B-Instruct. It utilizes a multi-stage reinforcement learning framework to facilitate structured and interpretable reasoning across eight representative aerial tasks. UAV-VL-R1 shows strong generalization and efficient deployment potential in aerial visual reasoning scenarios.

\subsection{Aerial Image Dataset}
Developing robust VLMs hinges not only on architectural advancements but also on access to high-quality, task-aligned datasets. While numerous aerial datasets exist, their task focus, supervision modality, and data granularity vary considerably. Table \ref{tab:aerial_datasets} summarizes a comprehensive comparison of representative aerial datasets.

\begin{table}[htbp]
\caption{Comparison of Representative Aerial Image Datasets}
\label{tab:aerial_datasets}
\centering
\begin{tabular}{lccc}
\toprule
\textbf{Dataset} & \textbf{Size} & \textbf{Task Domain} & \textbf{VQA Support} \\
\midrule
VisDrone~\cite{VisDrone}    & 10K+    & Detection, tracking        & \ding{55} \\
HazyDet~\cite{HazyDet}      & 383K+   & Detection in haze          & \ding{55} \\
UAV3D~\cite{UAV3D}          & 500K+   & 3D perception              & \ding{55} \\
UAVid~\cite{UAVid}          & 300+    & Semantic segmentation      & \ding{55} \\
SynDrone~\cite{SynDrone}    & 70K+    & Scene understanding        & \ding{55} \\
SkyScenes~\cite{SkyScenes}  & 33K+    & Scene understanding        & \ding{55} \\
FloodNet~\cite{FloodNet}    & 36K+    & VQA for disasters          & \ding{51} \\
RSVQA~\cite{RSVQA}          & 1066K+  & Remote sensing VQA         & \ding{51} \\
HRVQA~\cite{hrvqa}          & 1070K+  & General aerial VQA         & \ding{51} \\
\bottomrule
\end{tabular}
\end{table}

\begin{figure*}[!htp]  
\centering
\includegraphics[width=0.85\textwidth]{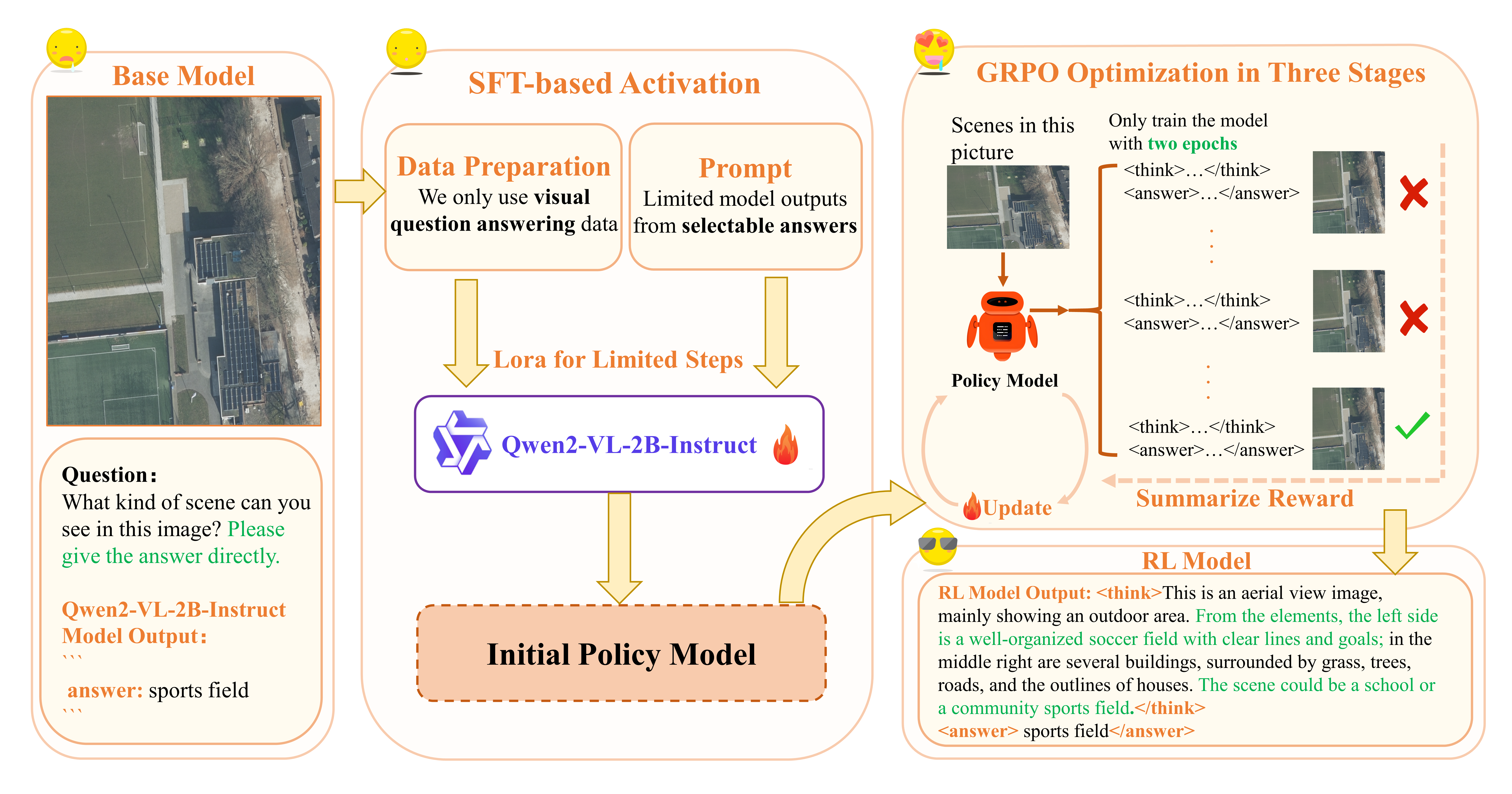}
\caption{Overview of the UAV-VL-R1 training framework. The model is first initialized via supervised fine-tuning (SFT) using prompt-constrained VQA data and LoRA-based lightweight adaptation. It is then progressively optimized across three GRPO-based reinforcement learning stages. The output follows a structured format composed of an interpretable reasoning trace \textit{\textless think\textgreater{}...\textless/think\textgreater{}} and a final prediction \textit{\textless answer\textgreater{}...\textless/answer\textgreater{}}.
}
\label{fig:Structure}
\end{figure*}
As shown, most aerial datasets (e.g., VisDrone, HazyDet, UAV3D, UAVid) \cite{VisDrone,HazyDet,UAV3D,UAVid} focus on conventional vision tasks such as object detection or segmentation, without language supervision. While a few datasets (e.g., FloodNet, RSVQA, HRVQA) \cite{FloodNet,RSVQA,hrvqa} support visual question answering, they still present critical limitations: FloodNet has a limited sample size and focuses on post-disaster assessment; RSVQA data is mainly based on remote sensing satellite images, which differ significantly from aerial images in terms of semantic content; HRVQA, as the largest aerial VQA dataset (1070K), was constructed using a semi-artificial synthesis method, resulting in annotation errors and language redundancy, making it difficult to use directly for high-quality training and evaluation of VLM.
To address the above issues, we construct HRVQA-VL, a refined and VLM-compatible subset of HRVQA consisting of 50{,}019 carefully filtered samples. The dataset covers eight representative reasoning tasks. All samples follow a structured ``image-question-answer'' format, supporting explicit reasoning output and reinforcement learning training process. HRVQA-VL offers a stable evaluation benchmark for multi-task generalization in aerial VQA and fills the gap for structured reasoning in UAV perception tasks.

\subsection{Exploration of Reinforcement Learning in Visual Reasoning}
In recent years, RL has demonstrated strong potential in improving the reasoning capabilities of Large Language Models (LLMs). Techniques like Direct Preference Optimization (DPO) \cite{dpo} and Proximal Policy Optimization (PPO) \cite{ppo}, widely adopted in the GPT series \cite{GPT-4}, have significantly enhanced logical consistency and output stability by designing reward functions that guide models to generate structured and coherent output \cite{SFT-or-RL}. This reward-driven training paradigm also holds promise for VLMs, encouraging interpretable reasoning beyond pattern imitation \cite{Reason-RFT,OpenVLThinker}.
However, conventional RL methods like PPO often suffer from high variance and unstable policy updates, particularly in complex structured tasks. To address this, we adopt Group Relative Policy Optimization (GRPO) \cite{grpo}, which estimates advantages based on intra-group reward differences. GRPO improves training stability and reduces variance, making it especially suited for structured reasoning tasks \cite{G1,RLHF-V}. Based on this, we propose UAV-VL-R1: a lightweight VLM tailored for UAV visual reasoning.

As illustrated in Fig.~\ref{fig:Structure}, the framework follows a two-stage training strategy. During the first stage, the model undergoes SFT using prompt-aligned, visual question answering data. This step leverages LoRA-based parameter-efficient tuning~\cite{LoRA} to establish an initial policy that aligns visual inputs with semantically grounded textual outputs. In the second stage, the model is refined via a three-phase reinforcement learning process based on Group Relative Policy Optimization (GRPO). GRPO encourages the exploration of diverse reasoning paths by comparing intra-group outputs and rewarding higher-quality samples, enabling the model to learn structured, consistent, and explainable inference trajectories. The final model produces outputs in a dual-tag format: a \textit{\textless think\textgreater{}...\textless/think\textgreater{}} trace that captures the reasoning process, and a \textit{\textless answer\textgreater{}...\textless/answer\textgreater{}} token representing the predicted answer.

\section{Method}
This section presents the detailed architecture, training procedure, and optimization pipeline of UAV-VL-R1. Building upon insights from post-training strategies in open-source language models \cite{TULU-3}, particularly those involving multi-stage optimization and reinforcement learning without human preference labels—we introduce a novel training framework that combines Supervised Fine-Tuning (SFT) with multi-stage reinforcement learning via Group Relative Policy Optimization (GRPO). GRPO serves as the core mechanism, enabling the model to construct structured and interpretable reasoning paths without depending on preference-based reward models. By integrating structural constraints with reward-driven learning, UAV-VL-R1 demonstrates strong generalization and reasoning robustness across UAV-based visual tasks, establishing a practical and scalable post-training paradigm for vision-language models.

\subsection{UAV-VL-R1 Training and Reasoning Pipeline}
\begin{figure*}[!ht]
\centering
\includegraphics[width=\textwidth]{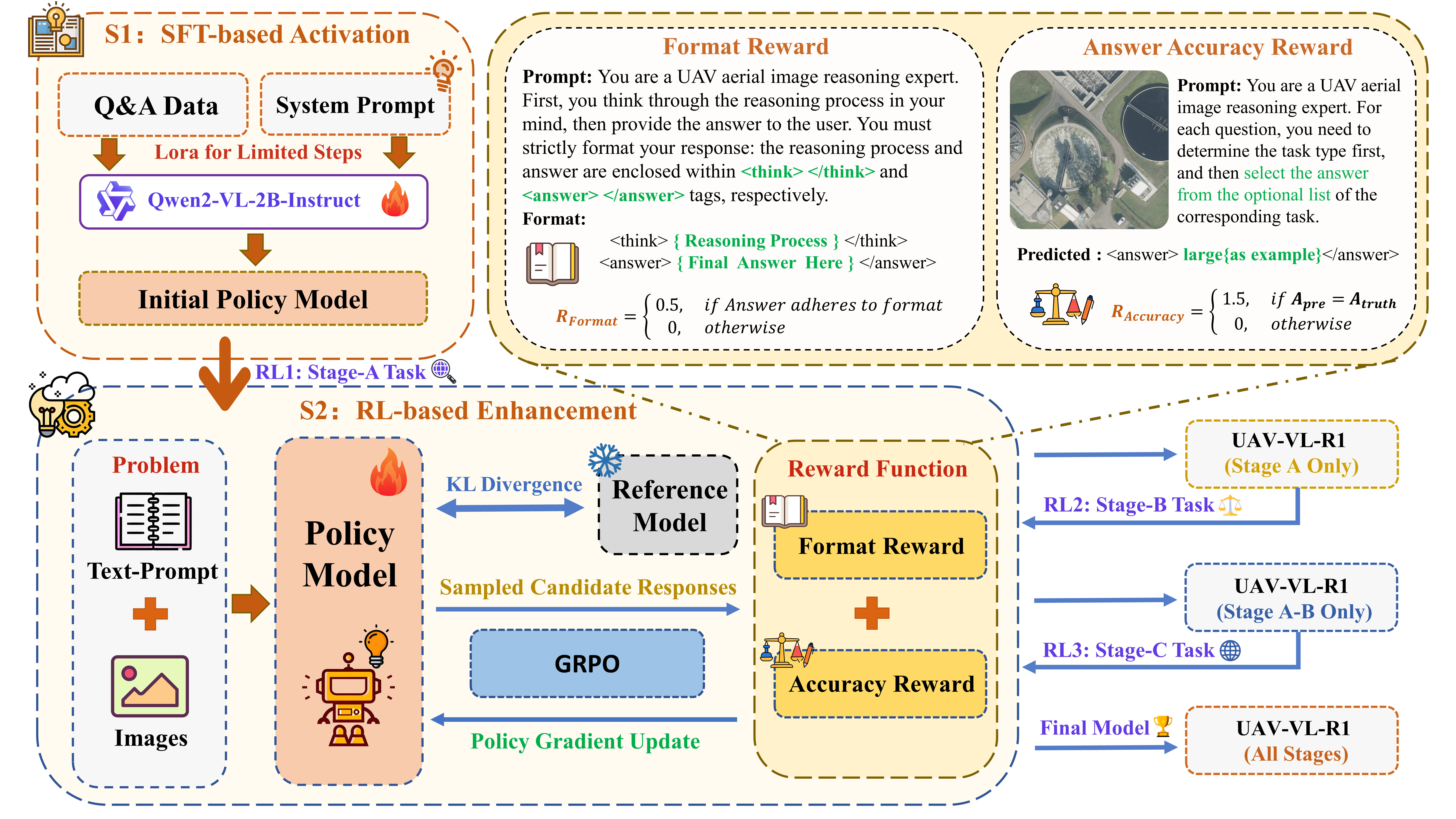}
\caption{Multi-stage visual reasoning framework of UAV-VL-R1. The pipeline consists of four components: SFT for semantic initialization, a reward function module for structural guidance, a GRPO module for policy refinement and reasoning path modeling, and a multi-stage training schedule (RL1–RL3) organized by task complexity. The model takes aerial images and natural language questions as input and produces structured outputs, including a reasoning trace wrapped in \textit{\textless think\textgreater{}...\textless/think\textgreater{}} and a final answer enclosed in \textit{\textless answer\textgreater{}...\textless/answer\textgreater{}}.}
\label{fig:framework}
\end{figure*}

As illustrated in Fig.~\ref{fig:framework}, the UAV-VL-R1 training and reasoning framework consists of four core components: a Supervised Fine-Tuning (SFT) module, a reward function module, a GRPO-based reinforcement learning module, and a multi-stage task training pipeline. The input comprises high-resolution aerial images and Q\&A data. The output is a structured reasoning process and a final answer, both formatted explicitly.
The training follows a staged curriculum. First, the SFT module performs semantic alignment and initializes the model’s language reasoning capacity using lightweight Lora-based tuning. The model then enters three successive reinforcement learning stages (RL1 to RL3), each optimized using Group Relative Policy Optimization (GRPO). To enhance the structured reasoning of the model, two types of reward functions are applied together during reinforcement training: format reward and accuracy reward. Each RL stage corresponds to a progressively more complex task set. RL1 focuses on basic attribute reasoning (e.g., color, size, binary questions); RL2 includes mid-level reasoning tasks (e.g., object counting, transportation classification, shape recognition); RL3 addresses high-level spatial and semantic understanding (e.g., location relationships and scene comprehension). This staged design facilitates gradual acquisition of structured reasoning, enhancing training stability and cross-task generalization. It also establishes the groundwork for the subsequent experiments on transfer and generalization.
\subsection{SFT-based Reasoning Activation}
\label{subsec:sft}
In the initial stage, we apply Supervised Fine-Tuning (SFT) \cite{SupervisedCL,VisualIT} to the base model Qwen2-VL-2B-Instruct to establish fundamental multimodal input--output mappings. To reduce training cost and improve adaptation to aerial image tasks, we adopt Low-Rank Adaptation (LoRA) \cite{LoRA} with lora-rank=32 and scaling factor lora-alpha=48. The backbone model is frozen, while the visual encoder remains trainable to adapt to the UAV-specific semantic space. This process activates the model’s contextual reasoning potential and provides a stable initial policy distribution for the subsequent GRPO optimization phase\cite{SFT-RL}.The SFT stage uses high-quality visual question-answering annotated samples from the HRVQA-VL dataset we constructed. Each sample is denoted as $(i,q,r,a)$, where $p$ is the input image, $q$ is the question, $r$ is the step-by-step reasoning path, and $a$ is the final answer. The training objective is to maximize the conditional likelihood of producing the reasoning and answer sequence $(r, a)$, given the image-question pair $(i, q)$:
\begin{equation}
\mathcal{L}_{\text{SFT}} = -\mathbb{E}_{(i,q,r,a)\sim\mathcal{D}}\left[\sum_{t=1}^{T}\log \pi_\theta(y_t \mid i,q,y_{<t})\right]
\label{eq:sft_loss}
\end{equation}
where $\mathcal{D}$ is the dataset, $y$ the concatenated sequence of $r$ and a, and $\pi_\theta$ represents the model’s token distribution. The output model is denoted by  $\pi_{SFT}$, which serves as the initialization for the next stage, ensuring a robust foundation for reinforcement learning.

\subsection{RL-based Reasoning Enhancement}
\label{subsec:RL-based}
\subsubsection{Overview}After SFT, the model proceeds to the reinforcement learning stage to improve its structured reasoning capabilities and generalization across tasks. We employ Group Relative Policy Optimization (GRPO)  instead of traditional Proximal Policy Optimization (PPO). Unlike PPO, which relies heavily on a learned value function and incurs high computational overhead, GRPO eliminates the need for value function approximation by estimating advantages directly through relative rewards among candidate responses. This makes GRPO particularly suitable for structured and rule-constrained visual reasoning tasks.

\subsubsection{Group Relative Policy Optimization Definition}Formally, let $P(\mathcal{Q})$ denote the distribution over training questions, and $q \in \mathcal{Q}$ be a sampled question. GRPO samples a set of candidate outputs $\{o_1,o_2,\cdots,o_G\}$ from the old policy $\pi_{\theta_{\mathrm{old}}}$. The current policy model $\pi_\theta$ is then optimized via the following objective function: 

\begin{equation}
\begin{aligned}
\mathcal{J}_{GRPO}(\theta) 
&= \mathbb{E}\bigl[q \sim P(Q), \{o_i\}_{i=1}^G \sim \pi_{\theta_{\text{old}}}(O|q)\bigr] \\
&\frac{1}{G} \sum_{i=1}^G \frac{1}{|o_i|} \sum_{t=1}^{|o_i|} 
\min\Biggl[
\frac{\pi_{\theta}(o_i|q)}{\pi_{\theta_{\text{old}}}(o_i|q)} A_i, \\
&\operatorname{clip}\left(
\frac{\pi_{\theta}(o_i|q)}{\pi_{\theta_{\text{old}}}(o_i|q)},\, 1{-}\varepsilon,\, 1{+}\varepsilon
\right)
\Biggr] - \beta D_{KL}[\pi_{\theta} \| \pi_{\text{ref}}]
\end{aligned}
\end{equation}

\begin{equation}
\mathbb{D}_{KL}[\pi_\theta||\pi_{ref}]=\frac{\pi_{ref}(o_i|q)}{\pi_\theta(o_i|q)}-\log\frac{\pi_{ref}(o_i|q)}{\pi_\theta(o_i|q)}-1
\end{equation}
where $\frac{\pi_\theta\left(o_i\middle| q\right)}{\pi_{\theta_{old}}\left(o_i\middle| q\right)}$ is the policy ratio, which controls the update step size. Hyperparameters $\varepsilon$ and $\mathrm\beta$ define clipping thresholds and regularization strength. The KL divergence\cite{ppo,KL-divergence,High-dimensional} term $\mathbb{D}_{KL}[\pi_\theta||\pi_{ref}]$ ensures that $\pi_\theta$ does not deviate excessively from the reference model $\pi_{ref}$. Specifically, for each input problem q, we sample a set of candidate outputs $\left\{o_1,o_2,\cdots,o_G\right\}$ from the old strategy. For each candidate response $o_i$, a scalar reward $\ r_1$ is computed based on rule-based criteria. The relative advantage\cite{DeepSeek-R1} $A_{i}$ is calculated by standardizing the rewards within the group:
\begin{equation}
A_i=\frac{r_i-\max(\{r_1,r_2,...,r_G\})}{\operatorname{std}(\{r_1,r_2,...,r_G\})}
\end{equation}
In this formula, mean$\left(\left\{r_1,r_2,\ldots,r_G\right\}\right)$ represents the average value of all rewards in the group, and std $\left(\left\{r_1,r_2,\ldots,r_G\right\}\right)$ represents the standard deviation. This standardization process can effectively hedge reward shifts between samples, improving the robustness of strategy gradient estimation and training stability. All rewards are computed from deterministic functions—no external reward model is needed. The reward function of this strategy incorporates structural format restriction reward and answer correctness reward, as detailed in Section~\ref{subsec:reward_design}.

\subsubsection{Details}Part S2 of Fig.~\ref{fig:framework} shows that the GRPO-based reinforcement learning stage is divided into three progressive task phases (RL1 to RL3). Each phase introduces increasingly complex reasoning tasks and uses GRPO to update the model policy iteratively. Specifically, a set of candidate responses is sampled from the previous policy $\pi_{\theta_{\text{old}}}$ to form a response group for a given input question. Each response is evaluated with rule-based rewards, considering structural format compliance and answer correctness. The rewards are then normalized within the group to compute the relative advantage for each response, which is used to update the current policy $\pi_{\theta}$. 
A KL divergence regularization term is also added to constrain policy drift and stabilize training. This design of sampling-based intra-group comparison, reward normalization, and advantage-driven optimization mitigates the instability and high cost associated with value-function-based reinforcement learning. The model gradually builds robust reasoning capabilities through multi-phase optimization, ranging from attribute-level recognition to spatial semantic modeling.

\subsection{Reward Design for Visual Reasoning Tasks}
\label{subsec:reward_design}
To guide the model in producing structurally coherent and semantically accurate reasoning outputs, we adopt a rule-based reward strategy inspired by prior works \cite{grpo,VLM-R1}. Specifically, we introduce two types of rewards: Format Reward and Accuracy Reward.

\subsubsection{Format Reward}The format reward evaluates whether the model's output adheres to a predefined structured format. The expected response should contain both of the following segments: \textit{\textless think\textgreater{}...reasoning steps...\textless/think\textgreater{}, \textless answer\textgreater{}...final answer...\textless/answer\textgreater{}}. If both tags are correctly included in the response, the reward is $0.5$; otherwise, the reward is $0$. This design encourages the model to generate outputs in a structured manner, facilitating better interpretability and easier extraction of reasoning chains.

\subsubsection{Accuracy Reward}The accuracy reward measures whether the content within the \textit{\textless answer\textgreater{}...\textless/answer\textgreater{}} tag matches the ground-truth answer. If correct, the reward is $1.5$; otherwise, $0$. This component directly evaluates the semantic correctness of the model's final decision.The total reward for each training sample is defined as:
\begin{equation}
r(a) = r_{\mathrm{format}}(a) + r_{\mathrm{accuracy}}(a)
\label{eq:total_reward}
\end{equation}
where: $r_{\mathrm{format}}(a) \in \{0,0.5\}$, $r_{\mathrm{accuracy}}(a) \in \{0,1.5\}$. As a result, the total reward $r(a) \in [0,2]$, with $2.0$ indicating the optimal case where both the structural format and semantic correctness are achieved.

\section{Experiment \& Results}
\label{sec:experiments}
This section introduces the experimental setup, details of its implementation, and evaluation results of UAV-VL-R1. We aim to systematically validate its generalization ability, interpretability, and deployment potential in UAV-based aerial visual reasoning tasks.

\subsection{Experimental Setup}
\label{subsec:setup}
\subsubsection{Dataset}We constructed the HRVQA-VL dataset based on the HRVQA \cite{hrvqa} dataset for VLMs training and evaluation. The dataset comprises 50,019 visual question answering samples covering eight representative UAV reasoning tasks: \textit{color, size, yes/no, shape, number, transportation, location, and scene}. We split the dataset into a training set and a test set with an 85:15 ratio, resulting in 42,465 training samples and 7,554 test samples. We further divided the training samples into 19,187 samples for the SFT stage and 23,278 samples for the three stages of post-training reinforcement learning. Specifically, we allocated 5,434 samples to Stage-A, 9,257 samples to Stage-B, and 8,587 samples to Stage-C.

\subsubsection{Implementation Details}We conduct all experiments on a hybrid hardware setup that includes 2$\times$ NVIDIA A100 (40GB PCIe) and 2$\times$ RTX A6000 (48GB) GPUs. We adopt PyTorch \cite{PyTorch}, DeepSpeed \cite{DeepSpeed}, and FlashAttention-2 \cite{FlashAttention} for efficient single-node multi-GPU distributed training. We initialize from the Qwen2-VL-2B-Instruct \cite{Qwen2-VL} as the base model. For the SFT stage, we apply LoRA fine-tuning \cite{LoRA} using an early stopping strategy \cite{fine-plm}, with a batch size of 1, gradient accumulation steps of 4, and mixed precision with bfloat16. The model is trained for 4 epochs. In the reinforcement learning stage, the model employs the multi-stage optimization strategy (RL1-RL3) for the GRPO algorithm described in section~\ref{subsec:RL-based}. Each RL stage inherits weights from the previous one and is trained with batch size $= 1$, gradient accumulation $= 2$, bfloat16 precision, and 2 epochs per stage. This stage-wise training strategy progressively activates the model's structured reasoning capabilities and enhances its stability in cross-task generalization.

\subsubsection{Evaluation Strategies}We design three evaluation strategies on the HRVQA-VL dataset to assess the generalization and reasoning capabilities of UAV-VL-R1: 
\begin{enumerate}
    \item \textbf{Cross-Task Generalization}: The model is trained on a subset of tasks and zero-shot evaluated on unseen task groups. We compare its transferability against general-purpose VLMs;
    \item \textbf{Multi-Task Generalization}: The fully trained UAV-VL-R1 is evaluated on the full test set containing all eight tasks. We benchmark its aggregated performance and examine whether general VLMs can maintain high output quality under format-constrained reasoning, or if their lack of internal reasoning paths leads to performance degradation;
    \item \textbf{Ablation Study}: We analyze the impact of SFT initialization on RL training stability and final accuracy by comparing two paths: ``GRPO-only'' and ``SFT + GRPO''.
\end{enumerate}

\subsubsection{Baselines and Evaluation Metrics}We consider three categories of baselines: 
\begin{enumerate}
    \item \textbf{General-purpose VLMs:} InstructBLIP~\cite{InstructBLIP}, Qwen2-VL-Instruct (2B/7B/72B)~\cite{Qwen-VL,Qwen2-VL,Qwen2-report}.
    \item \textbf{SFT Variants:} Qwen2-VL-2B-Instruct with supervised fine-tuning.
    \item \textbf{RL Variants:} Qwen2-VL-2B-Instruct further optimized via a three-stage RL strategy.
\end{enumerate}
The evaluation metric is answer accuracy. All models are required to output in the following structure:\textit{\textless think\textgreater{}...reasoning steps...\textless/think\textgreater{}, \textless answer\textgreater{}...final answer...\textless/answer\textgreater{}}. We extract the content within the \textit{\textless answer\textgreater{}} tag for evaluation and compare it with ground-truth labels. Accuracy is computed as:
\begin{equation}
\mathrm{Accuracy} = 100 \times \frac{1}{N} \sum_{i=1}^{N} \hat{f}\left(\hat{y}_o = y_i\right)
\end{equation}
where $N$ is the total number of test samples, $\hat{y}_o$ is the predicted answer, $y_i$ is the ground-truth label, and $\hat{f}$ is the indicator function that returns 1 if $\hat{y}_o = y_i$, and 0 otherwise.

\subsubsection{Fairness Disclaimer}This study focuses on structured reasoning tasks under aerial VQA scenarios. It is important to clarify that task formulation, dataset structure, and evaluation format factors influence the performance differences observed between general-purpose VLMs and our task-specific model. Therefore, the results presented here demonstrate how the model performs under these specific conditions, and we should not interpret them as a general ranking of the model’s capability or overall utility in wider multimodal settings.

\subsection{Cross-Task Generalization}
\label{sec:Cross-Task}
We design a staged experiment to systematically assess the cross-task transfer and generalization abilities of UAV-VL-R1 in aerial visual reasoning. 
\subsubsection{Task Grouping and Evaluation Protocol}We divide the eight tasks into three disjoint groups and apply GRPO optimization on the SFT-initialized Qwen2-VL-2B-Instruct model by training it only on one group. The model is then evaluated in a zero-shot manner on the remaining two unseen groups to measure transferability and structural reasoning generalization.The task groups are defined as follows (consistent with Fig. \ref{fig:Data_inform}).
\begin{enumerate}
    \item \textbf{Stage A:} Basic attribute recognition (color, size, yes/no);
    \item \textbf{Stage B:} Object identification and counting (number, shape, transportation);
    \item \textbf{Stage C:} Spatial reasoning and scene understanding (location, scene).
\end{enumerate}
All models are evaluated under a structured prompting format, requiring outputs in the following structure:\textit{\textless think\textgreater{}...reasoning steps...\textless/think\textgreater{}, \textless answer\textgreater{}...final answer...\textless/answer\textgreater{}}.

\subsubsection{Results on cross-task generalization} As shown in Table~\ref{tab:cross_task}, UAV-VL-R1 trained only on Stage-A achieves 48.37\% and 46.59\% accuracy on unseen Stage-B and Stage-C tasks, respectively, with an average score of 60.44\%, outperforming all baseline models, including the 72B-parameter Qwen2-VL-Instruct. This demonstrates the effectiveness of GRPO in enabling strong cross-task transfer in a lightweight architecture.
\begin{table*}[t]
\caption{Cross-Task Generalization Performance of General VLMs and Single-stage RL Post-Training Models. Each row corresponds to a model under evaluation, while each column reports its zero-shot accuracy on a specific task stage. ``Overall'' denotes the average accuracy across all three stages.}
\label{tab:cross_task}
\centering
\begin{tabular}{lcccc}
\toprule
\textbf{Model} & \textbf{Stage-A} & \textbf{Stage-B} & \textbf{Stage-C} & \textbf{Overall} \\
\midrule
Qwen2-VL-2B-Instruct (Base)       & 35.16 & 21.96 & 14.76 & 23.96 \\
InstructBLIP-7B                   & 38.77 & 23.21 & 30.50 & 30.82 \\
Qwen2-VL-7B-Instruct              & 49.95 & 34.38 & 18.92 & 34.41 \\
Qwen2-VL-72B-Instruct             & 56.82 & 51.39 & 40.05 & 49.42 \\
Qwen2-VL-2B-Instruct (SFT)        & 35.82 & 30.61 & 20.98 & 29.13 \\
UAV-VL-R1 (Stage A only)          & \textbf{86.38} & 48.37 & 46.59 & 60.44 \\
UAV-VL-R1 (Stage B only)          & 64.01 & \textbf{62.71} & 25.82 & 50.84 \\
UAV-VL-R1 (Stage C only)          & 61.66 & 38.06 & \textbf{49.37} & 49.69 \\
UAV-VL-R1 (Ours, all stages)      & 86.27 & 63.40 & 64.22 & \textbf{71.30} \\
\bottomrule
\end{tabular}
\end{table*}
Further analysis reveals that Stage-A contains low-level attribute recognition tasks that facilitate visual-language alignment, helping the model generalize better across task types. In contrast, Stage-C enhances structured reasoning for spatial understanding but exhibits weaker transfer to lower-level perceptual tasks such as color or size. Overall, UAV-VL-R1 shows robust transferability even when trained on a single task group. These results validate the efficacy of GRPO in guiding structured reasoning and support the feasibility of multi-stage reinforcement learning in multi-task VLM training.

\subsection{Generalized Multi-Task Reasoning and Structured Output Adaptability Evaluation}
To comprehensively evaluate the performance of UAV-VL-R1 on multi-task aerial image reasoning, we design a unified zero-shot evaluation experiment to benchmark it against a series of mainstream VLMs. This experiment focuses on two key questions.
\begin{enumerate}
    \item How well UAV-VL-R1 performs in terms of generalization across eight aerial reasoning tasks;
    \item Analyze the impact of structured output prompts on the performance of different models to evaluate their adaptability to structured reasoning format constraints.
\end{enumerate}
\begin{table}[h]
\caption{Evaluation of reasoning accuracy under two prompting strategies on the HRVQA-VL full test set. Each row represents a model under evaluation. ``Zero-shot Plain'' indicates free-form generation without structural guidance, while ``Zero-shot Prompting'' enforces explicit reasoning and answer formatting using predefined \textit{\textless think\textgreater{}} and \textit{\textless answer\textgreater{}} tags.}
\label{tab:multi_task}
\scriptsize
\centering
\begin{tabular}{lcc}
\toprule
\textbf{Model} & \textbf{Zero-shot Plain (\%)} & \textbf{Zero-shot Prompting (\%)} \\
\midrule
Qwen2-VL-2B-Instruct           & 36.60 & 22.20 \\
Qwen2-VL-2B-Instruct (SFT)     & 52.60 & 26.93 \\
InstructBLIP-7B                & 37.67 & 32.91 \\
Qwen2-VL-7B-Instruct           & 43.83 & 38.40 \\
Qwen2-VL-72B-Instruct          & 51.46 & 46.67 \\
UAV-VL-R1 (Ours, all stages)   & \textbf{68.94} & \textbf{72.13} \\
\bottomrule
\end{tabular}
\end{table}
\subsubsection{Baselines and Evaluation Strategies}We consider InstructBLIP-7B and the Qwen2-VL series (2B/7B/72B) as representative general-purpose VLMs. For comparison, we also include the SFT version of Qwen2-VL-2B-Instruct as a pre-RL baseline. 
All models are evaluated on the full HRVQA-VL test set in a zero-shot setting, using two input prompting strategies.
\begin{enumerate}
    \item \textbf{Zero-shot Plain:} No structural reasoning template is provided; models generate answers freely.
    \item \textbf{Zero-shot Prompting:} Structured templates are added to the input to enforce explicit reasoning output in the format: \textit{\textless think\textgreater{}...reasoning steps...\textless/think\textgreater{}, \textless answer\textgreater{}...final answer...\textless/answer\textgreater{}}.
\end{enumerate}

\subsubsection{Results Analysis}As shown in Table~\ref{tab:multi_task}, the UAV-VL-R1 model achieves the highest accuracy (68.94\%) in the plain setting. It significantly outperforms all baseline models, including the 72B-parameter Qwen2-VL-72B-Instruct model. Notably, its performance further improves to 72.13\% in the Prompting setting, demonstrating not only strong generalization but also robust structured reasoning capability under format-constrained conditions.

In contrast, most general-purpose VLMs exhibit performance drops under structured prompting. For example, Qwen2-VL-2B-Instruct drops from 36.60\% to 22.20\%, and Qwen2-VL-7B-Instruct falls from 43.83\% to 38.40\%. Even the largest model, Qwen2-VL-72B-Instruct, shows a decrease from 51.46\% to 46.67\%. These results suggest that without specific training on structured reasoning formats, general-purpose VLMs struggle to adapt, leading to a significant degradation in answer quality.

Notably, the UAV-VL-R1 model not only outperforms all baselines in both settings but also benefits from the structured prompting, turning the formatting constraints into effective signals for reasoning enhancement. This shows that the model has internalized its structural reasoning ability through joint SFT initialization and GRPO-based reinforcement learning.

In summary, this experiment confirms that UAV-VL-R1 exhibits superior multi-task generalization and robust adaptation to structured output formats, outperforming large-scale general models even under challenging constraints. It also highlights a critical limitation in current general-purpose VLMs: despite having large parameter sizes and good zero-shot capabilities, they still lack the ability to consistently generate structured reasoning outputs without dedicated training mechanisms.

\begin{table*}[bt]
\caption{Ablation study evaluating the effect of SFT on the generalization performance of GRPO. Each row corresponds to a specific training checkpoint, defined by its training route (with or without SFT) and the completed reinforcement learning stages. ``Stage-A,'' ``Stage-B,'' and ``Stage-C'' represent evaluation accuracy on three task groups in the HRVQA-VL test set. ``Overall'' denotes the average score across all three groups. Route-1 indicates direct GRPO training without SFT, while Route-2 includes SFT initialization prior to GRPO.}
\label{tab:ablation_study}
\centering
\begin{tabular}{lllcccc}
\toprule
\textbf{Model Variant} & \textbf{Training Route} & \textbf{Trained Stages} & \textbf{Stage-A (\%)} & \textbf{Stage-B (\%)} & \textbf{Stage-C (\%)} & \textbf{Overall (\%)} \\
\midrule
GRPO-A                & Route-1 (no SFT)         & Only Stage A            & 73.39 & 30.05 & 21.74 & 41.72 \\
SFT-GRPO-A            & Route-2 (with SFT)       & Only Stage A            & \textbf{86.38} & \textbf{48.37} & \textbf{46.59} & \textbf{60.45} \\
GRPO-AB               & Route-1                  & Stage A + B             & 71.27 & \textbf{65.37} & 19.24 & 51.96 \\
SFT-GRPO-AB           & Route-2                  & Stage A + B             & \textbf{86.82} & 61.52 & \textbf{39.25} & \textbf{62.53} \\
GRPO-ABC              & Route-1                  & All Stages              & 70.17 & \textbf{64.94} & 43.32 & 59.48 \\
SFT-GRPO-ABC (Ours)   & Route-2                  & All Stages              & \textbf{86.27} & 63.40 & \textbf{64.22} & \textbf{71.30} \\
\bottomrule
\end{tabular}
\end{table*}

\subsection{Ablation Study: The Role of Supervised Fine-Tuning in Multi-stage Reinforcement Learning Training}
To systematically examine the role of SFT in multi-stage reinforcement learning training, we design an ablation study to analyze its impact on model generalization and training stability. 

\subsubsection{Design of Training Paths and Evaluation Stages}We compare the following two training routes:
\begin{itemize}
    \item \textbf{Route-1: GRPO-direct} – Directly applying GRPO across all three training stages on the base model without any prior SFT.
    \item \textbf{Route-2: SFT-initialized GRPO} – First, SFT is applied to the base model for Lora fine-tuning training, and then GRPO is applied across the same task stages sequentially.
\end{itemize}
Each training route is decomposed into three stage-specific tasks: Stage-A, Stage-B, and Stage-C. The formulation of each task aligns with the definitions provided in Fig.\ref{fig:Data_inform}. Three model checkpoints are evaluated for each route: trained on Stage-A only, Stage-A + B, and all stages (A + B + C). After each stage, the model is evaluated on the test sets of the three stages to assess its overall generalization performance.

\subsubsection{Results on Ablation Study} As shown in Table~\ref{tab:ablation_study}, we observe that in the early stage (Stage-A only), SFT-GRPO-A achieves an average accuracy of 60.45\%, nearly 19 points higher than GRPO-A (41.72\%).
This confirms that SFT provides a stable initial policy distribution, accelerating GRPO convergence and enhancing generalization even with limited training scope. However, as training progresses to Stage-B, GRPO-AB slightly outperforms SFT-GRPO-AB on Stage-B tasks (65.37\% vs. 61.52\%), and this pattern persists in Stage-B accuracy even after full-stage training (64.94\% vs. 63.40\% for GRPO-ABC vs. SFT-GRPO-ABC). This suggests that while SFT improves alignment and early stability, it may limit the model’s exploration capability, particularly for tasks involving numeric or compositional reasoning.

This observation aligns with prior findings~\cite{SFT-or-RL}, indicating that reasoning skills are often emergent through reinforcement, while SFT tends to introduce imitative but rigid reasoning structures. Nonetheless, SFT-GRPO-ABC achieves the highest overall accuracy of 71.30\%, outperforming GRPO-ABC (59.48\%). The improvement is especially significant on Stage-C spatial and semantic tasks (64.22\% vs. 43.32\%), highlighting that SFT contributes to more stable and explainable reasoning in complex visual scenes by enhancing language-vision alignment.

In summary, these results demonstrate that SFT plays a critical supporting role in multi-stage GRPO optimization. It improves early-stage convergence, enhances generalization, and strengthens structural reasoning for semantically rich tasks. However, the interaction between SFT and RL is non-additive. There is a trade-off between semantic consistency and policy diversity. This prompts the exploration of optimal combinations under varying task types, data complexity, and task ordering in future studies.

\subsection{Training Dynamics and Reward Progression}
\label{subsec:training_dynamics}
To better understand the learning dynamics in our \textsc{grpo}-based multi-stage training framework, we analyze three key metrics during the final RL-three stage: training loss, \textit{KL Divergence}, and two types of reward signals—\textit{format reward} and \textit{accuracy reward}. These metrics help reveal how the model converges and how policy shifts occur.

\begin{figure}[!htp]  
\centering
\includegraphics[width=\linewidth]{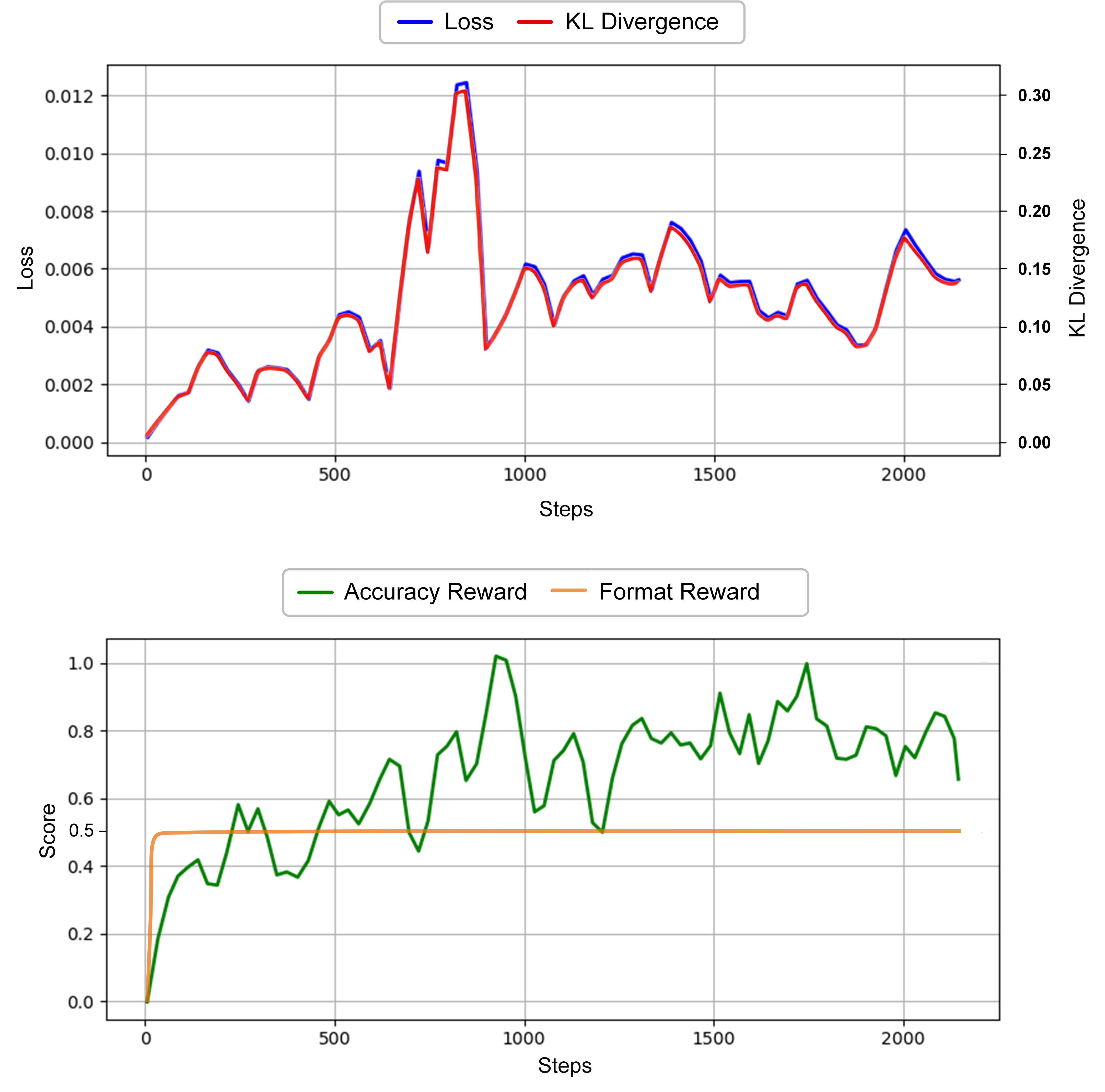}  
\caption{Training loss, KL divergence, format reward, and accuracy reward curves during the final RL-3 stage.}
\label{fig:guan4}
\end{figure}
As shown in Figure~\ref{fig:guan4}, the training loss steadily decreases and begins to stabilize around step 1400, indicating that the model is approaching convergence. The \textit{KL Divergence}, plotted alongside the loss curve on a separate y-axis, exhibits a nearly identical trajectory, suggesting similar trends in optimization behavior.
The lower part of Figure~\ref{fig:guan4} shows the trajectories of the two \textsc{grpo} reward components. The format reward rises sharply during the first few steps and quickly approaches its upper bound of 0.5 within approximately 30 steps, indicating that the model rapidly learns to produce syntactically structured outputs. In contrast, the accuracy reward remains low during the initial phase and begins to rise around step 200—an ``aha'' moment \cite{VLM-R1} that reflects the model's acquisition of task-specific semantic reasoning. The reward then gradually increases and stabilizes around step 1400, in line with the loss trend.

Furthermore, we observe that fluctuations in \textit{KL Divergence} exhibit an inverse correlation with changes in accuracy reward: When KL values rise, accuracy reward tends to decline, vice versa. This behavior reflects the typical role of KL-based regularization in reinforcement learning—balancing policy exploration and stability.
Together, these observations highlight a two-phase learning pattern: rapid structural alignment followed by gradual semantic reasoning refinement, both governed by the dynamics of GRPO training.

\section{Conclusion}
We present UAV-VL-R1, a lightweight vision-language model tailored for structured reasoning in UAV-based aerial imagery. To address conventional VLMs’ limited generalization and poor interpretability in aerial tasks, we propose a unified training framework that combines SFT with multi-stage reinforcement learning. Central to this framework is GRPO, which, combined with a rule-based reward function, reduces policy variance and enhances stability, guiding the model to generate structured and semantically consistent reasoning paths.

To support training and evaluation, we construct HRVQA-VL, a cleaned and structured dataset covering eight representative UAV tasks, ranging from basic attribute recognition to spatial reasoning. Through LoRA-based SFT and GRPO optimization, UAV-VL-R1 effectively mitigates issues such as sparse rewards and unstable convergence, while enhancing cross-task generalization.
Experiments demonstrate that UAV-VL-R1 outperforms robust VLM baselines, including Qwen2-VL-72B, despite having only 2 billion parameters. It also demonstrates strong deployment potential, requiring just 3.9GB in FP16 inference and compressible to 2.5GB via INT8 quantization. This makes it suitable for real-world UAV scenarios like disaster monitoring and low-altitude perception.

In future work, we aim to expand the dataset scope, incorporate more task types and modalities, and further refine the reward design to improve RL stability. We are deeply committed to contributing to this field and hope that UAV-VL-R1 can serve as a solid foundation for advancing the development of scalable visual reasoning systems in aerial intelligence applications.

\vfill
\bibliographystyle{ieeetr}
\bibliography{Mybib}
\end{document}